# Detection of the most influential variables for preventing postpartum urinary incontinence using machine learning techniques

José Alberto Benítez-Andrades[1] , María Teresa García-Ordás[2] ,
María Álvarez-González[3], Raquel Leirós-Rodríguez[4] and
Ana F López Rodríguez[3]

## Abstract

**Background:** Postpartum urinary incontinence is a fairly widespread health problem in today's society among women who have given birth. Recent studies analysing the different variables that may be related to Postpartum urinary incontinence have brought to light some variables that may be related to Postpartum urinary incontinence in order to try to prevent it. However, no studies have been found that analyse some of the intrinsic and extrinsic variables of patients during pregnancy that could give rise to this pathology.

**Objective:** The objective of this study is to assess the most influential variables in Postpartum urinary incontinence by means of machine learning techniques, starting from a group of intrinsic variables, another group of extrinsic variables and a mixed group that combines both types.

**Methods:** Information was collected on 93 patients, pregnant women who gave birth. Experiments were conducted using different machine learning classification techniques combined with oversampling techniques to predict four variables: urinary incontinence, urinary incontinence frequency, urinary incontinence intensity and stress urinary incontinence.

**Results:** The results showed that the most accurate predictive models were those trained with extrinsic variables, obtaining accuracy values of 70% for urinary incontinence, 77% for urinary incontinence frequency, 71% for urinary incontinence intensity and 93% for stress urinary incontinence.

**Conclusions:** This research has shown that extrinsic variables are more important than intrinsic variables in predicting problems related to postpartum urinary incontinence. Therefore, although not conclusive, it opens a line of research that could confirm that the prevention of Postpartum urinary incontinence could be achieved by following healthy habits in pregnant women.

## Introduction and related work

The International Continence Society defines urinary incontinence (UI) as any involuntary loss of urine.[1] Three types of UI are distinguished: (i) stress, in which the leakage is caused by jumping, sneezing, coughing, etc., that is, by an increase in intra-abdominal pressure not compensated by the muscular activity of the perineum; (ii) urgency (or bladder hyperactivity), in which the leakage is caused by anarchic contractions of the detrusor muscle of the

[1]SALBIS Research Group, Department of Electric, Systems and Automatics Engineering, Universidad de León, León, Spain
[2]SECOMUCI Research Group, Escuela de Ingenierías Industrial e Informática, Universidad de León, León, Spain
[3]Faculty of Health Sciences, Universidad de León, Ponferrada, Spain
[4]SALBIS Research Group, Nursing and Physical Therapy Department, Universidad de León, Ponferrada, Spain

**Corresponding author:**
María Teresa García-Ordás, SECOMUCI Research Group, Escuela de Ingenierías Industrial e Informática, Universidad de León, Campus de Vegazana s/n, C.P. 24071 León, Spain.
Email: mgaro@unileon.es

bladder that provoke an imperious and sudden desire to urinate; and (iii) mixed, when the patient presents a combined clinical picture of the two aforementioned ones.[2]

Regardless of the type of UI, it is a major medical, social and economic problem, which is increasingly prevalent and greatly reduces the quality of life of women who suffer from it.[3,4] Postpartum urinary incontinence is a very common health problem affecting a large number of women worldwide and it is one of the most common postpartum complications.[5] There are a large number of research studies and systematic reviews that provide different conclusions regarding the relationship between the different variables and the occurrence of this symptom of urinary incontinence.[6,7]

All of these studies yield results that may be useful from a traditional statistical point of view. They have shown that some of the recorded risk variables that influence this health problem are: vaginal delivery, advanced age at gestation, advanced maternal Body Mass Index (BMI), excess weight gain during pregnancy, diabetes, episiotomy, forceps delivery or gestational and prenatal UI.[5] However, as a result of the different artificial intelligence techniques, such as machine learning, among others, they are being increasingly used in this field and, thanks to them, results are being obtained that are of interest to the scientific community and that represent an advance and a novelty in the automation of diagnoses and in the prevention of possible health problems.[8–11]

For example, in the Kasasbeh et al.[12] study, different machine learning techniques (bagging, AdaBoost and logistic regression, among others) were used to identify or predict the factors that influence different postpartum complications.

In the study carried out by Venkatesh et al.[13], different predictive machine learning models (extreme gradient boosting model, lasso regression model, logistic regression) are used to predict a woman's risk of postpartum haemorrhage on admission at delivery.

Other researchers such as Betts et al.[10] succeeded in developing models for predicting neonatal outcomes using machine learning. Their model achieved a high discrimination for neonatal respiratory distress syndrome, in particular, reinforcing tree models gave the best results on the datasets they worked with.

Zhang et al.[8] focused on developing and validating a machine learning model with the aim of predicting the risk of postpartum depression in pregnant women. They used two datasets containing data from 15,197 women and 53,972 women, respectively. They trained five different machine learning models, random forest, decision tree, extreme gradient boosting (XGboost), regularized logistic regression, and multilayer perceptron (MLP).

Machine learning models were also used to identify the pregnant women's risk of persistent post-traumatic stress following labour.[9] The authors of this study were able to create models that predicted post-traumatic labor stress by 75.8%. This was done using decision trees.

Despite studies applying artificial intelligence techniques to different postpartum health problems, no studies have been found that analyse the possible relationship between intrinsic and extrinsic variables with four variables: occurrence of UI and stress UI and frequency and intensity of UI.

This research therefore studies the application of different machine learning models, adjusting the hyperparameters and adding oversampling techniques, on six different sets of variables to predict the four outcome variables related to UI.

This paper is organized as follows: In section "Methodology," the classic methods used to compare results and the architectures used in our proposal are explained. Experiments and results, as well as the dataset used, are detailed in section "Experiments and results." Finally, we conclude in section "Discussion and conclusions."

## Methodology

### Classic methods

In order to see which variables are more influential in postpartum urinary incontinence, different variations of the following machine learning techniques have been used: Gaussian Naive Bayes, Complement Naive Bayes, K-nearest neighbors (KNN), and decision trees. Experiments have also been carried out using oversampling techniques such as RandomOver and Smote. In this section, the vanilla version of the methods used will be introduced.

*Gaussian Naive Bayes.* The Gaussian Naive Bayes classifier is based on the Naive Bayes algorithm and the Naive Bayes algorithm is based on the Bayes Theorem used to calculate conditional probabilities.

$P(A)$ being the probability of event $A$; $P(B)$ the probability of event $B$; $P(A|B)$ the probability of event $A$ given $B$; $P(B|A)$ the probability of event $B$ given $A$ and $P(A \cap B)$ the probability of both event $A$ and event $B$ occurring, then the Bayes Theorem for conditional the calculation of probabilities is given by the following equation.

$$P(A|B) = \frac{P(B|A) \times P(A)}{P(B)} \qquad (1)$$

If we use the classic Naive Bayes classifier, we have to take into account that the variables have to be totally independent of each other. On the other hand, Naive Bayes has the advantage that it is a simple classifier and requires little training data to achieve a good result.

In our case, the Gaussian Naive Bayes is used. This modification of the initial algorithm allows the work to be carried out with continuous data assuming that they follow a normal (Gaussian) distribution while still maintaining the advantages of the classic method. The Gaussian Naive Bayes conditional probability is calculated by terms of the following equation

$$P(x) = \frac{1}{\sqrt{2\pi\sigma^2}} e^{-\frac{(x-\mu)^2}{2\sigma^2}} \quad (2)$$

where $\mu$ is the mean, calculated by the following equation

$$\mu = \frac{1}{n} \sum_{i=1}^{n} x_i \quad (3)$$

and $\sigma$ is the standard deviation calculated by the following equation

$$\sigma = \sqrt{\frac{1}{n-1} \sum_{i=1}^{n} (x_i - \mu)^2} \quad (4)$$

*Complement Naive Bayes.* This algorithm, derived from the previous ones, is usually used when the dataset with which we are working has unbalanced data, that is, many more elements belonging to some classes than to others.

To solve the problem of unbalanced classes, this algorithm differs from Naive Bayes, because it calculates the probability that an element has of belonging to all classes, instead of calculating the probability that the element has of belonging to a specific class.

The process is as follows: First, for each class, the probability that the data does not belong to it, is calculated. Then, the smallest value of those obtained is selected. And finally, this lowest probability is selected, which tells us that this class is not the one to which the element belongs. This way of calculating which class the data belongs to, is why the method is called Complement Naive Bayes.

*K-nearest neighbors.* KNN is a well-known method in which the class of the input data is calculated by taking into account the distances from that data to the rest of the elements of each class. More specifically, it is based on selecting the $K$ closest neighbors to the input data. The input data will be classified within the class to which the largest number of the selected $k$ neighbors belong.

The distance between the input data and the rest of the data in the dataset is usually calculated using the Euclidean distance, given by the following equation

$$d(x, x') = \sqrt{(x_1 - x'_1)^2 + \cdots + (x_n - x'_n)^2} \quad (5)$$

After calculating the distances, the input $x$ is assigned to the class with the highest probability (6).

$$P(y = j | X = x) = \frac{1}{K} \sum_{i \in A} I(y^i = j) \quad (6)$$

Choosing the best value for parameter k depends on the type of problem, the type of data, and other factors, so it is important to make a grid search to select the most appropriate value for the problem.

*Decision tree.* The trees are made up of the root node that represents all the data and is the node that will be divided to give rise to the decision nodes. The leaf nodes are those in which we can no longer do more splits. The decision tree algorithm takes into account two important concepts in order to split the data: Entropy and Gini index. Entropy can be defined as the amount of information needed to describe an element and it is given by the following equation

$$E = -\sum_{i=1}^{n} p_i * log(p_i) \quad (7)$$

The Gini index gives us an idea of how homogeneous the data is. An index of 0 means that the sample is totally homogeneous and an index of 1 means that the sample is totally heterogeneous. This value is given by the following equation

$$G = 1 - \sum_{i=1}^{n} p_i^2 \quad (8)$$

The algorithm consists of the following steps. In the first place, the root node is selected by taking into account those data that generate less entropy and more information. Then, at each iteration, the entropy and information gain are calculated for each unused node. The base node that generates the least entropy and the greatest gain in information is selected. Splits are made to generate data subsets, and this process is repeated for each subset until the complete tree is created.

*Oversampling.* Oversampling techniques are widely used on unbalanced data sets to improve methods and results.[14,15] Ways of balancing classes can be to delete data from the majority class, or duplicate data from the minority class. In this article we will use RandomOver and SMOTE as oversampling methods.

- RandomOver: It is a very simple technique that randomly selects samples from the minority class with a replacement and duplicates them, until both sets are balanced.
- SMOTE: Using this technique,[16] no new duplicate data is generated, but synthetic data similar to that of the minority class is created. In the data generation process, first, a sample is randomly selected to generate similar data. The k closest neighbors to that random sample are calculated, which, by default, are five neighbors. The new data are generated using a distance metric to compute the difference between the feature vector and the selected neighbors. That difference vector is multiplied by a random value of between 0 and 1 and this new vector is the new generated data.

Table 1. Description of the extrinsic variables collected from each patient and possible values.

| Variable | Description | Values |
| --- | --- | --- |
| AGE | Patient's age | Continuous |
| NUM_LABOURS | Number of labors | Continuous |
| DIC_NULLIPAROUS | Number of labors (dichotomous) | 0 = No previous labors, 1 = With previous labors |
| HEIGHT | Patient's height | Continuous (cm) |
| WEIGHT | Patient's weight | Continuous (kg) |
| BMI | Patient's BMI | Continuous |
| CAT_BMI | Patient's BMI category | 0 = Underweight, 1 = Normal-weight, 2 = Overweight |
| EXTRA_KG | Kg gained during pregnancy | Continous (kg) |
| CAT_EXTRAKG | Category according to kg gained during pregnancy | 0 = 10 kg or less, 1 = 11 to 15, 2 = 16 to 20, 3 = 21 to 25 |
| LABOUR_PREP | Variable describing how the preparation for labor went | 0 = Without help, 1 = With help |
| PROF_CHBPR | Professional who assisted in the preparation of the labor | 0 = No, 1 = Midwife, 2 = Midwife and Physiotherapy |
| PA_PREV | Previous physical activity undertaken | 0 = No, 1 = Yes |
| FREQ_PAPREV | Frequency of previous physical activity undertaken | 0 = No, 1 = 1 to 3 times a week, 2 = more than 3 times a week |
| IPAQ | Patient's International Physical Activity Questionnaires (IPAQ) score | 0 = Low, 1 = Moderate, 2 = Vigorous |
| WALKING | Variable on whether or not the patient walked during the pregnancy | 0 = No, 1 = Yes |
| STRENGTH | Whether the patient was strength training or not. | 0 = No, 1 = Yes |
| PILATES | Whether the patient was pilates training or not. | 0 = No, 1 = Yes |
| AQUAGYM | Whether the patient was aquagym training or not. | 0 = No, 1 = Yes |
| NUM_PA | Number of physical activities carried out | Continuous |

## Experiments and results

### Dataset

The dataset was obtained after a non-randomized controlled trial among women selected for their interest in participating in the study through information provided at their primary care centre (at their first obstetric consultation with the midwife and/or gynaecologist or through information leaflets, handed out at the care centre). Recruitment was carried out in three primary care centres attached to the same hospital, where all participants gave birth. During the year prior to this study (2019), 397 babies were born in the municipality in which this study was carried out.

The following inclusion criteria were defined for participation in the study (a) women between 18 and 40 years of age; (b) term delivery (week 37 or more); (c) singleton gestation and cephalic presentation; (d) pregnancy without complications or added risks during gestation for the development of a normal vaginal delivery; (e) no participation in any other psychoprophylaxis intervention; (f) delivery at the Hospital Nuestra Señora de Sonsoles (Spain). The following exclusion criteria were defined simultaneously (a) any contraindication to vaginal delivery; (b) medical diagnosis of any urogynecological pathology prior to the gestational process; (c) any history of caesarean delivery and/or history of perineal injury (episiomoty and/or perineal tear of any grade); (d) failure to give informed consent to participate in the study or failure to attend all scheduled intervention and/or assessment sessions.

Table 2. Description of the intrinsic variables collected from each patient and possible values.

| Variable | Description | Values |
| --- | --- | --- |
| WEEK_LABOUR | Week of labor | Continuous |
| INJURY | Was the patient injured? | 0 = No, 1 = Yes |
| EPISIOTOMY |  | 0 = No, 1 = Yes |
| TEARING | Did the patient have a tear? | 0 = No, 1 = Slight, 2 = Moderate |
| DURATION | Duration of labor | Continous (hours) |
| LITOTHOMY |  | 0 = No, 1 = Yes |
| POSTURE | Patient's posture during labor | 0 = Lithotomy, 1 = Side, 2 = Sitting / squatting, 3 = Standing |
| ANALGESIA | Did the patient have analgesia? | 0 = No, 1 = Yes |
| TYPE_ANALGESIA | Type of Analgesia | 0 = No, 1 = Local, 2 = Epidural, 3 = Espinal |
| TYPE_LABOUR | Type of labor and the need for the use of assistive devices | 0 = Euthocic, 1 = Forceps/Spatulae, 2 = Vacuum cups |
| KRISTELLER |  | 0 = No, 1 = Yes |
| WEIGHT_BABY | Weight of the baby | Continuous (g) |

Table 3. Description of the results variables collected from each patient and possible values.

| Variable | Description | Values |
| --- | --- | --- |
| VAS_PERINE | Intensity of pain in the perineum in the 6th week postpartum according to the Visual Analogue Scale | Continuous |
| UI | Did the patient have urinary incontinence? | 0 = No, 1 = Yes |
| FREQ_UI | Frequency of urinary incontinence | 0 = No, 1 = Sporadic, 2 = Daily |
| INT_UI | Intensity of urinary incontinence | 0 = No, 1 = Mild, 2 = Moderate, 3 = Severe |
| AFFECT_UI |  | 0 = No, 1 = Yes |
| BLADD_HYPER | Did the patient have bladder hyperactivity | 0 = No, 1 = Yes |
| UI_STRESS |  | 0 = No, 1 = Yes |
| UI_PREV | Did you previously have urinary incontinence? | 0 = No, 1 = bladder hyperactivity, 2 = stress |

Table 4. Selection of the best extrinsic, intrinsic, and global variables for each of the four outcome variables using SelectKBest.

| Variable | Input | SelectKBest results |
|---|---|---|
| UI | Intrinsic best | KRISTELLER, DIC_NULLIPAROUS, NUM_LABOURS |
| | Extrinsic best | AQUAGYM, GROUP, WEIGHT |
| | Best of all | AQUAGYM [ex], KRISTELLER [in], DIC_NULLIPAROUS [in] |
| Stress UI | Intrinsic best | KRISTELLER, DIC_NULLIPAROUS, NUM_LABOURS |
| | Extrinsic best | AQUAGYM, FREQ_PAPREV, IPAQ |
| | Best of all | AQUAGYM [ex], KRISTELLER [in], DIC_NULLIPAROUS [in] |
| UI Frequency | Intrinsic best | TYPE_PARTO, EPISIOTOM, DIC_NULLIPAROUS |
| | Extrinsic best | AQUAGY, STRENGTH, PILATES |
| | Best of all | TYPE_PARTO [in], AQUAGYM [ex], STRENGTH [ex] |
| UI Intensity | Intrinsic best | KRISTELLER, DIC_NULLIPAROUS, NUM_LABOURS |
| | Extrinsic best | AQUAGYM, GROUP, WEIGHT |
| | Best of all | AQUAGYM [ex], KRISTELLER [in], DIC_NULLIPAROUS [in] |

Table 5. Nomenclature and description of models trained for the experiments.

| Model | Description |
|---|---|
| GaussianNB | Gaussian Naive Bayes model with a default hyperparameter setting. |
| ComplementNB | Complement Naive Bayes model with a default hyperparameter setting. |
| KNN | k-Nearest Neighbors (KNN) model with a default hyperparameter setting. |
| DT | Decision Tree model with a default hyperparameter setting. |
| KNN improved | KNN model applying optimal hyperparameters according to an optimal parameter search performed using GridSearchCV. |
| DT improved | DT model applying optimal hyperparameters according to an optimal parameter search performed using GridSearchCV. |
| KNN imp.randover | KNN model by applying optimal hyperparameters and adding oversampling using the RandomOver technique. |
| KNN imp.SMOTE | KNN model by applying optimal hyperparameters and adding oversampling using the SMOTE technique. |

A dataset was generated by collecting information from 93 women who gave normal vaginal birth. Table 1 details all of the extrinsic variables, Table 2 details the intrinsic variables, and Table 3 sets out the result variables collected for each patient.

### Experimental setup

In the first stage, data pre-processing was carried out to clean the data and format them in a suitable format to work with. The variables were separated into intrinsic, extrinsic and outcome variables. Of the existing outcome

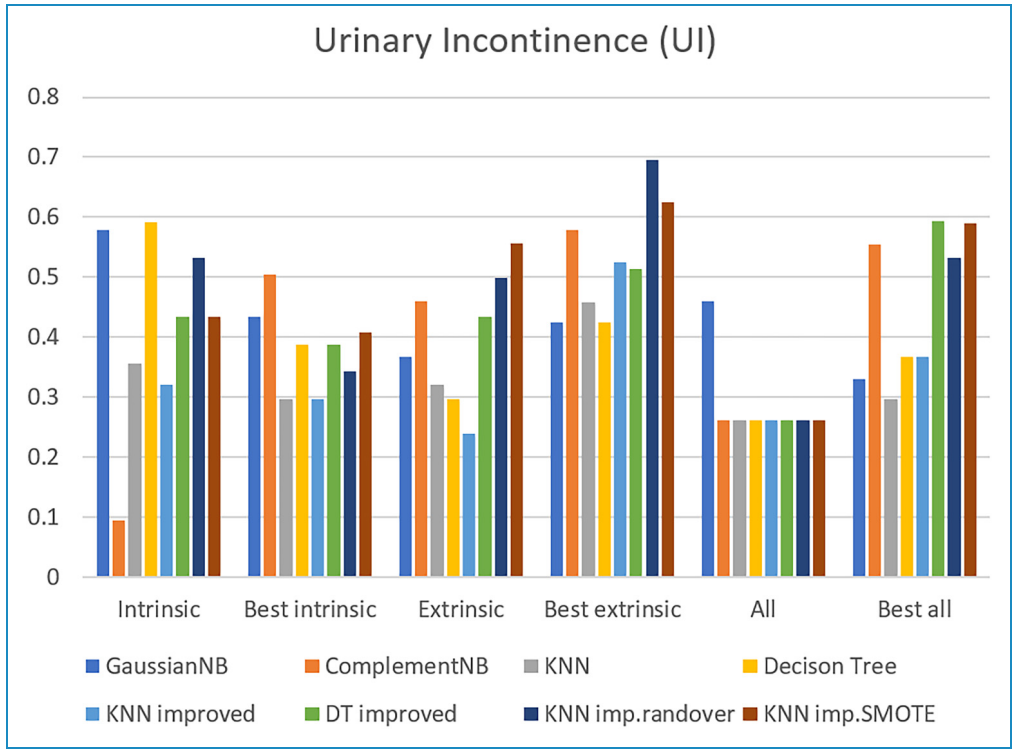

**Figure 1.** F1-score for the variable urinary incontinence (UI).

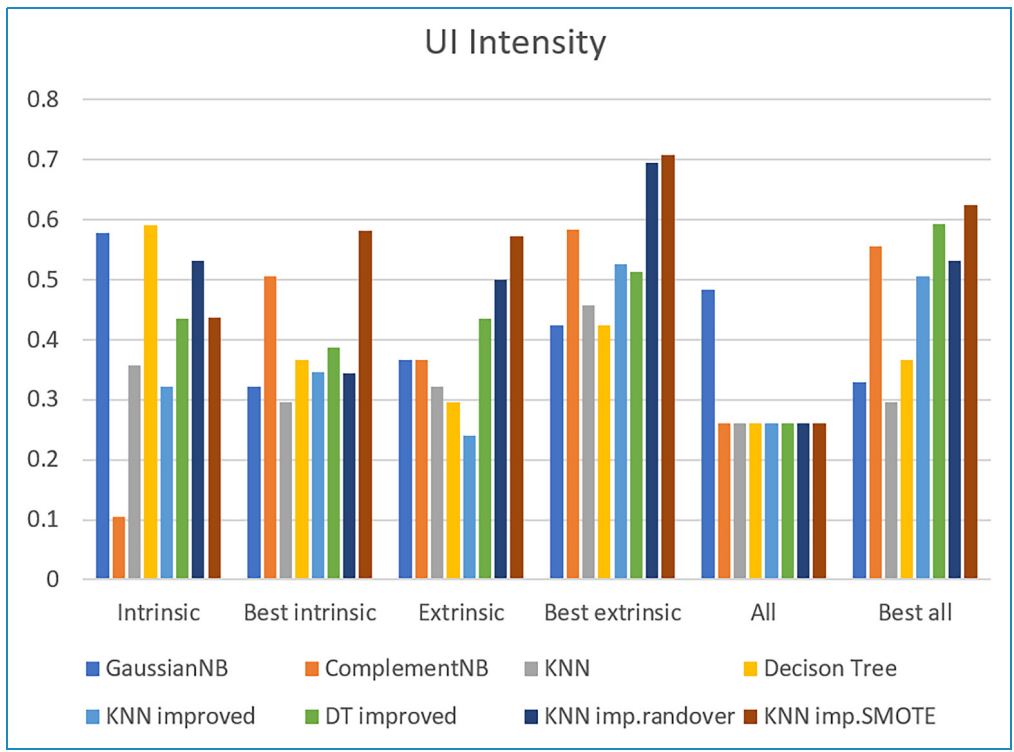

**Figure 2.** F1-score for the variable stress urinary incontinence (UI).

variables, four were found to be unbalanced with the sample of 93 patients collected and were therefore discarded for the present study. Thus, we worked with the following outcome variables:

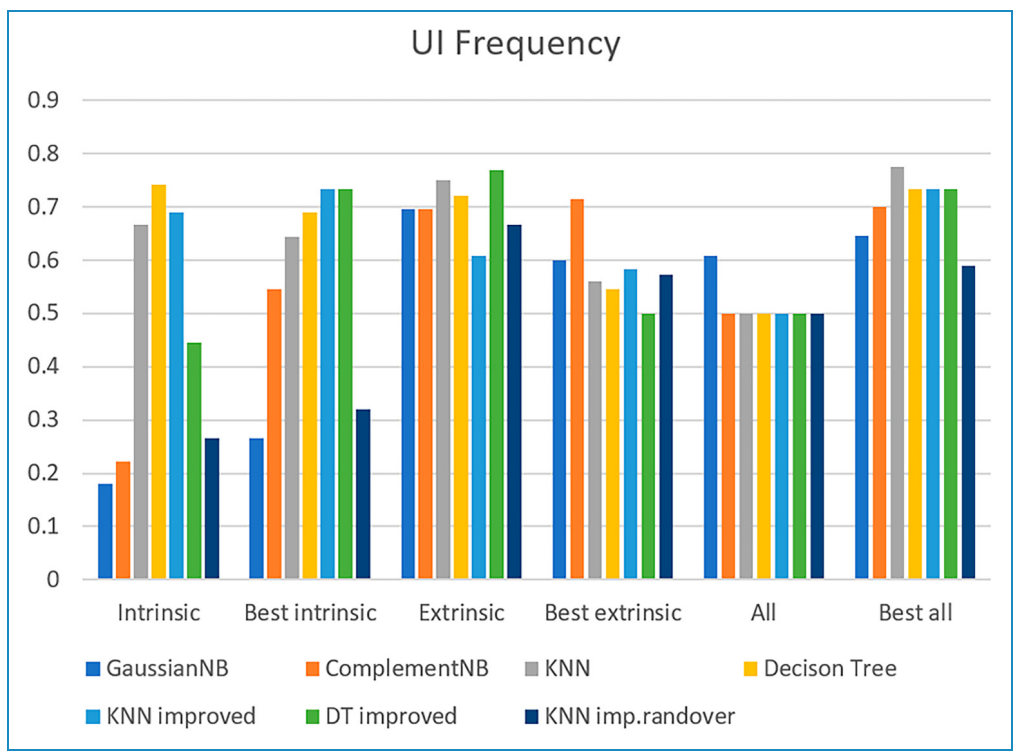

**Figure 3.** F1-score for the variable frequency of urinary incontinence (UI).

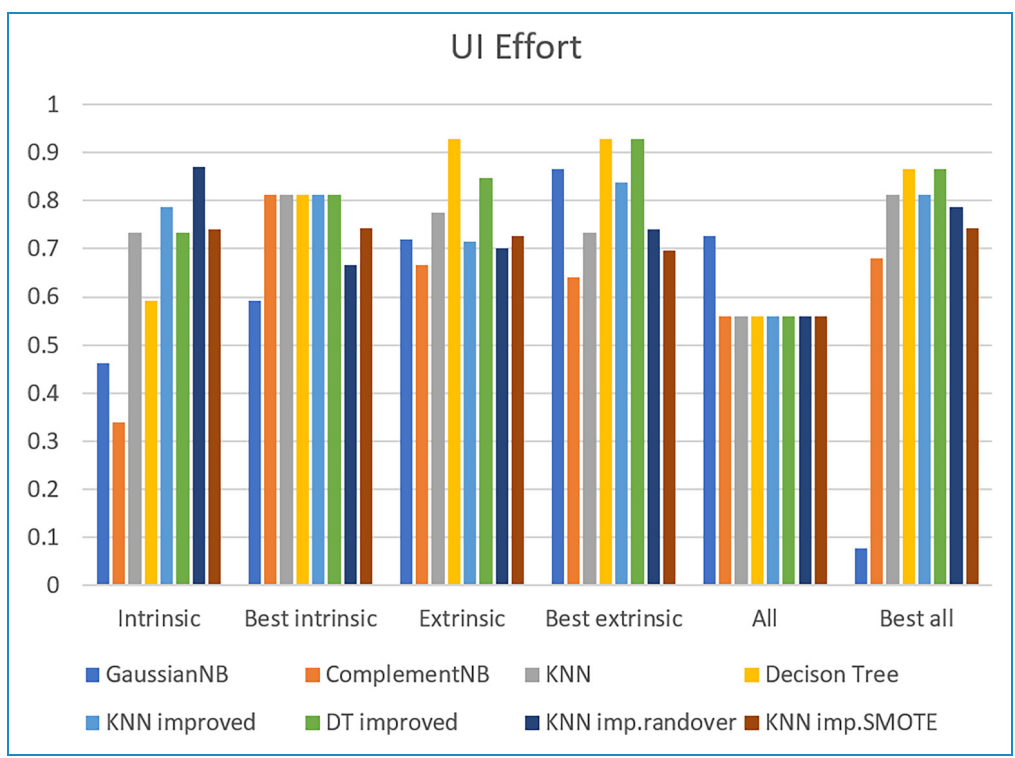

**Figure 4.** F1-score for the variable intensity of urinary incontinence (UI).

- **UI**: It represents whether or not the patient had urinary incontinence.
- **STRESS_UI**: It represents whether the urinary incontinence was provoked after exertion or not.
- **FREQ_UI**: It represents the frequency with which the patient experienced urinary incontinence.
- **INT_UI**: It represents the intensity of urinary incontinence experienced by each patient.

The group of intrinsic variables consists of 16 variables and the group of extrinsic variables consists of 15 variables. The variables belonging to each group are detailed in Table 1.

Pipelines were generated to convert the categorical variables using a LabelEncoder and the continuous variables were scaled. After this, a selection of best features was performed using SelectKBest and ExtraTrees. In this way, six groups of features were generated for each of the four outcome variables as follows:

- **Intrinsic variables:** This group was composed of all existing intrinsic variables. It is common to the experiments conducted with the four outcome variables.
- **Extrinsic variables:** This group was composed of all existing extrinsic variables. It is common to the experiments conducted with the four outcome variables.
- **All variables:** This group was composed of all existing intrinsic and extrinsic variables. It is common to the experiments conducted with the four outcome variables.
- **Best intrinsic variables**: For each outcome variable, a selection of the best intrinsic variables was made and a group of the three best performing intrinsic variables was created.
- **Best extrinsic variables**: For each outcome variable, a selection of the best extrinsic variables was made and a group of the three best performing extrinsic variables was created.
- **Best variables**: For each outcome variable, a selection of the best variables (both intrinsic and extrinsic) was made and a group of the three best performing variables was created.

Table 4 shows which variables formed part of the aforementioned groupings for each of the four variables resulting from this study.

A k-fold cross-validation with 10 folds has been performed for all experiments to avoid randomisation. Each model has been evaluated using the most appropriate hyper-parameter selection techniques according to the input data to which they were applied. In addition, RandomOver and SMOTE oversampling techniques were applied in some of them.

The experiments consisted of applying the following models to predict the four outcome variables (UI, FREQ_UI, INT_UI, and STRESS_UI). The nomenclature of the models used is detailed in Table 5.

## Results

Figures 1 and 2 show four graphs representing the results obtained by the eight aforementioned classification models, applied to each of the four outcome variables to be predicted using the six aforementioned groups of variables: presence of UI (Figure 1), frequency of UI (Figure 3), intensity of UI (Figure 4), and whether urinary incontinence was stress UI (Figure 2).

In addition, Table 6 shows the results of all of the experiments carried out.

By grouping the F1-score metrics obtained for each group of variables in all the experiments carried out, the means and standard deviations shown in Table 7 were obtained. The group of variables with the best mean F1-score metric was the group with the best extrinsic variables. This group consisted of three extrinsic variables that most influenced the models according to a previously run feature selector.

### Intrinsic versus Extrinsic

A two sample $t$-test was carried out to compare the F1-score in both the intrinsic and extrinsic variables.

There was a significant difference in F1-score between intrinsic ($M = 0.49$, SD $= 0.20$) and extrinsic ($M = 0.56$, SD $= 0.20$); $t(30) = -1.960$, $p = 0.045$).

### Intrinsic best versus Extrinsic best

A two-sample $t$-test was carried out to compare the F1-score in both the intrinsic best and extrinsic best groups of variables.

There was a significant difference in the F1-score between the intrinsic best ($M = 0.52$, SD $= 0.20$) and the extrinsic best ($M = 0.56$, SD $= 0.20$); $t(30) = -1.960$, $p = 0.002$).

### Best results

Table 8 shows the data for the three models with the best results for each of the four predicted variables. From among these data, it is possible to see the type of model applied, the input variables used and the F1-score value obtained.

## Ethical considerations

Approval for the study was obtained through the Ethics Committee of the University of Leon, Spain (code: ETICA-ULE-021-2018). All participants signed an informed consent form, in accordance with the Declaration of Helsinki (rev. 2013), and had the option to revoke their participation in the study at any time. Ethical regulations were respected as well as the Spanish Constitutional Law for the Protection of Data and for Biomedical Research in Human Participants.

## Discussion and conclusions

Following the experiments detailed in the previous section, the results obtained are discussed, highlighting that

Table 6. F1-score of the seven trained models for each of the four outcome variables.

| Predicted Var. | Classifier | F1-score | | | | | |
|---|---|---|---|---|---|---|---|
| | | Int. | Int. best | Ext. | Ext. best | All | Best of all |
| UI | GaussianNB | 0.58 | 0.43 | 0.37 | 0.42 | 0.46 | 0.33 |
| UI | ComplementNB | 0.10 | 0.50 | 0.46 | 0.58 | 0.26 | 0.55 |
| UI | KNN | 0.36 | 0.30 | 0.32 | 0.46 | 0.26 | 0.30 |
| UI | Decison tree | 0.59 | 0.39 | 0.30 | 0.42 | 0.26 | 0.37 |
| UI | KNN improved | 0.32 | 0.30 | 0.24 | 0.53 | 0.26 | 0.37 |
| UI | DT improved | 0.43 | 0.39 | 0.43 | 0.51 | 0.26 | 0.59 |
| UI | KNN imp. randover | 0.53 | 0.34 | 0.50 | **0.70** | 0.26 | 0.53 |
| UI | KNN imp. SMOTE | 0.43 | 0.41 | 0.56 | 0.62 | 0.26 | 0.59 |
| FREQ_UI | GaussianNB | 0.18 | 0.27 | 0.70 | 0.60 | 0.61 | 0.65 |
| FREQ_UI | ComplementNB | 0.22 | 0.55 | 0.70 | 0.71 | 0.50 | 0.70 |
| FREQ_UI | KNN | 0.67 | 0.64 | 0.75 | 0.56 | 0.50 | **0.77** |
| FREQ_UI | Decison tree | 0.74 | 0.69 | 0.72 | 0.55 | 0.50 | 0.73 |
| FREQ_UI | KNN improved | 0.69 | 0.73 | 0.61 | 0.58 | 0.50 | 0.73 |
| FREQ_UI | DT improved | 0.44 | 0.73 | **0.77** | 0.50 | 0.50 | 0.73 |
| FREQ_UI | KNN imp. randover | 0.27 | 0.32 | 0.67 | 0.57 | 0.50 | 0.59 |
| INT_UI | GaussianNB | 0.58 | 0.32 | 0.37 | 0.42 | 0.48 | 0.33 |
| INT_UI | ComplementNB | 0.11 | 0.50 | 0.37 | 0.58 | 0.26 | 0.55 |
| INT_UI | KNN | 0.36 | 0.30 | 0.32 | 0.46 | 0.26 | 0.30 |
| INT_UI | Decison tree | 0.59 | 0.37 | 0.30 | 0.42 | 0.26 | 0.37 |
| INT_UI | KNN improved | 0.32 | 0.34 | 0.24 | 0.53 | 0.26 | 0.50 |
| INT_UI | DT improved | 0.43 | 0.39 | 0.43 | 0.51 | 0.26 | 0.59 |
| INT_UI | KNN imp. randover | 0.53 | 0.34 | 0.50 | 0.70 | 0.26 | 0.53 |
| INT_UI | KNN imp. SMOTE | 0.44 | 0.58 | 0.57 | **0.71** | 0.26 | 0.62 |
| STRESS_UI | GaussianNB | 0.46 | 0.59 | 0.72 | 0.87 | 0.73 | 0.08 |
| STRESS_UI | ComplementNB | 0.34 | 0.81 | 0.67 | 0.64 | 0.56 | 0.68 |
| STRESS_UI | KNN | 0.73 | 0.81 | 0.77 | 0.73 | 0.56 | 0.81 |

(continued)

**Table 6.** Continued.

| Predicted Var. | Classifier | F1-score | | | | | |
|---|---|---|---|---|---|---|---|
| | | Int. | Int. best | Ext. | Ext. best | All | Best of all |
| STRESS_UI | Decison Tree | 0.59 | 0.81 | **0.93** | **0.93** | 0.56 | 0.87 |
| STRESS_UI | KNN improved | 0.79 | 0.81 | 0.71 | 0.84 | 0.56 | 0.81 |
| STRESS_UI | DT improved | 0.73 | 0.81 | 0.85 | **0.93** | 0.56 | 0.87 |
| STRESS_UI | KNN imp.randover | 0.87 | 0.67 | 0.70 | 0.74 | 0.56 | 0.79 |
| STRESS_UI | KNN imp.SMOTE | 0.74 | 0.74 | 0.73 | 0.70 | 0.56 | 0.74 |

**Table 7.** Mean F1-score and standard desviation (SD) obtained according to the group of variables used in all experiments.

| Variable group | N | Mean | SD |
|---|---|---|---|
| Intrinsic | 31 | 0.49 | 0.20 |
| Intrinsic best | 31 | 0.52 | 0.19 |
| Extrinsic | 31 | 0.56 | 0.20 |
| Estrinsic best | 31 | **0.61** | 0.15 |
| All | 31 | 0.42 | 0.15 |
| Best of all | 31 | 0.58 | 0.20 |

**Table 8.** Three best models for each variable.

| Variable | Input | Model | F1-score |
|---|---|---|---|
| UI | Extrinsic best | KNN imp. randover | 0.70 |
| | Extrinsic best | KNN imp. SMOTE | 0.62 |
| | Best of all | DT improved | 0.59 |
| Stress UI | All | DT | 0.93 |
| | Extrinsic best | DT | 0.93 |
| | Extrinsic best | DT improved | 0.93 |
| UI Frequency | Best of all | KNN | 0.77 |
| | Extrinsic | DT improved | 0.77 |
| | Extrinsic | KNN | 0.75 |
| UI Intensity | Extrinsic best | KNN imp. SMOTE | 0.71 |
| | Extrinsic best | KNN imp. randover | 0.70 |
| | Best of all | KNN imp. SMOTE | 0.62 |

extrinsic variables had the greatest influence on the four outcome variables analysed, that is, on the occurrence of postpartum urinary incontinence as well as on the frequency and intensity of UI and whether the source of this UI was due to biomechanical stress factors. Training the models with all intrinsic variables resulted in an average F1-score of 48.92% versus 55.65% using all extrinsic variables as input parameters of the models. Analysing the groups in which the three best intrinsic variables were selected, an average F1-score of 52.25% was obtained compared to a 61.35% success rate in the models using the three best extrinsic variables. The difference in both cases is statistically significant, so it is possible to determine that extrinsic variables have a greater influence on the occurrence, frequency and intensity of postpartum urinary incontinence.

These findings are consistent with previous research pointing to lifestyle-related and therefore modifiable risk factors such as obesity,[7] type of delivery,[17] or episiotomy.[18] These results, therefore, corroborate that postpartum urinary incontinence could be avoided if some lifestyle changes were made by the patients in their daily lives. It is especially relevant that the weight gained during pregnancy has shown significant results. Excessive weight gain during pregnancy increases the abdominal weight overload on the perineal musculature. This directly increases the pressure on these muscles.[19] But, in addition, the increased intra-abdominal pressure associated with overweight increases the mechanical load on this musculature indirectly.[20] The same physiological processes are probably responsible for the significant results found in relation to physical activity before and during pregnancy. Since the practice of physical activity would prevent excessive weight gain during pregnancy.[21] Furthermore, the inclusion in the model of all the activities practiced by the participating women supports the current recommendation on the importance of the intensity at which the exercise is carried out and not the particular discipline.[22] In parallel, the influence of the number of previous deliveries was also significant in the model generated. This phenomenon may be due to the fact that the accumulation of vaginal deliveries favors the degeneration of the muscular and fascial structures of support and continence.[17,23] Conversely, counselling of the pregnant woman by different health professionals and attendance at childbirth preparation classes would favor the maintenance of self-care and behavioral patterns that are preventive and/or non-injurious to the perineum.[24–26] The group of intrinsic variables (or not under the full control of the pregnant woman) includes obstetric characteristics traditionally associated with increased risk of UI: gestational age at delivery and weight of the baby (two related variables),[5] the presence of perineal tear or episiotomy[27] and the duration of labour.[28] The only intrinsic variable included and currently agreed to have no influence on the development of postpartum UI is the use of analgesia during labour.[29,30] In any case, our results do not question the possible structural impact of these events. Rather, they point to the lesser impact on the incidence of UI than the other group of variables analysed in this research. The main limitations of this study are the small sample size and that the dataset was obtained after a non-randomized controlled trial (level III of evidence). In addition, the amount of physical activity performed by the women participants was recorded through self-reported information rather than by objective quantitative methods such as pedometers or accelerometers. The authors recognize that these factors limit the generalizability of the results obtained. However, it opens up a line of research to carry out an intervention at a training level so that women implement a change of habits before or during pregnancy and confirm the findings found in our results.

**Acknowledgements:** We are grateful for the willingness of the participants in this study and the support provided by Lyu and Zaggy.

**Contributorship:** José Alberto Benítez-Andrades: Conceptualization, Data curation, Methodology, Software, Visualization, Validation, Writing-Original draft preparation. María-Teresa García-Ordás: Data curation, Writing-Original draft preparation. María Álvarez-González: Conceptualization, Supervision, Writing-Reviewing and Editing. Ana F. López Rodríguez: Conceptualization, Supervision, Writing-Reviewing and Editing. Raquel Leirós-Rodríguez: Data curation, Methodology, Software, Visualization, Validation, Writing-Reviewing and Editing.

**Declaration of Conflicting Interests:** The authors declare(s) that there is no conflict of interest.

**Ethical approval:** Approval for the study was obtained through the Ethics Committee of the University of Leon, Spain (code: ETICA-ULE-021-2018).

**Funding:** This research received no specific grant from any funding agency in the public, commercial, or not-for-profit sectors.

**Guarantor:** José Alberto Benítez-Andrades.

**ORCID iDs:** José Alberto Benítez-Andrades  https://orcid.org/0000-0002-4450-349X
María Teresa García-Ordás  https://orcid.org/0000-0002-3796-3949